\documentclass[twoside]{article}

\usepackage[accepted]{aistats2021}
\usepackage[utf8]{inputenc} % allow utf-8 input
\usepackage[T1]{fontenc}    % use 8-bit T1 fonts
\usepackage{url}            % simple URL typesetting
\usepackage{booktabs}
\usepackage{multirow}
\usepackage{graphicx}
\usepackage{subcaption}
\usepackage{floatrow}
\usepackage{amsfonts}       % blackboard math symbols
\usepackage{algorithm}
\usepackage{algpseudocode}
\usepackage{nicefrac}       % compact symbols for 1/2, etc.
\usepackage{microtype}      % microtypography
\usepackage{amsmath}
\usepackage{nccmath, mathtools}
\usepackage{enumitem}
\usepackage{floatrow}
\usepackage{wrapfig}
\usepackage{comment}
\usepackage[round]{natbib}
\usepackage[svgnames]{xcolor}
\definecolor{inkblue}{RGB}{25 25 112}
\usepackage[colorlinks=true,linkcolor=red,citecolor=inkblue]{hyperref}%

\newfloatcommand{capbtabbox}{table}[][\FBwidth]

\newcommand{\x}{\mathbf{x}}
\newcommand{\oo}{\mathbf{o}}
\newcommand{\uu}{\mathbf{u}}
\newcommand{\zz}{\mathbf{z}}
\newcommand{\mm}{\mathbf{m}}

\begin{document}



\twocolumn[

\aistatstitle{Variational Selective Autoencoder: Learning from Partially-Observed Heterogeneous Data}
% Variational Selective Autoencoders for Learning from Partially-Observed Data
\aistatsauthor{ Yu Gong\textsuperscript{1,2} \And 
                Hossein Hajimirsadeghi\textsuperscript{1} \And  
                Jiawei He\textsuperscript{1} \And
                Thibaut Durand\textsuperscript{1} \And
                Greg Mori\textsuperscript{1,2}}

\aistatsaddress{ \textsuperscript{1}Borealis AI \And  
                 \textsuperscript{2}Simon Fraser University  
                 } ]

\begin{abstract}
  Learning from heterogeneous data poses challenges such as combining data from various sources and of different types. 
Meanwhile, heterogeneous data are often associated with missingness in real-world applications due to heterogeneity and noise of input sources.
In this work, we propose the variational selective autoencoder (VSAE), a general framework to learn representations from partially-observed heterogeneous data. 
VSAE learns the latent dependencies in heterogeneous data by modeling the joint distribution of observed data, unobserved data, and the imputation mask which represents how the data are missing. 
It results in a unified model for various downstream tasks including data generation and imputation. Evaluation on both low-dimensional and high-dimensional heterogeneous datasets for these two tasks shows improvement over state-of-the-art models.
\end{abstract}

%   Learning from heterogeneous data poses challenges such as combining data from various sources and of different types. 
% Meanwhile, heterogeneous data are often associated with missingness in real-world applications due to heterogeneity and noise of input sources.
% In this work, we propose the variational selective autoencoder (VSAE), a general framework to learn representations from partially-observed heterogeneous data. 
% VSAE learns the latent dependencies in heterogeneous data by modeling the joint distribution of observed data, unobserved data, and the imputation mask which represents how the data are missing. 
% It results in a unified model for various downstream tasks including data generation and imputation. Evaluation on both low-dimensional and high-dimensional heterogeneous datasets for these two tasks shows improvement over state-of-the-art models.

\section{Introduction}
\label{sec:intro}

Learning from data is an integral part of artificial intelligence. 
A typical assumption of learning algorithms that the data is fully-observed is clearly unrealistic in many settings, therefore handling missing data has been a long-standing problem~\citep{NIPS1993_767, Schafer:1997}. The data generation process combined with the observation mechanism by which data are hidden makes learning from those data much more complex.
Moreover, conventional algorithms rely heavily on clean \textit{homogeneous} data, yet varied, \textit{heterogeneous} data are a common setting for learning. In fact, heterogeneity is ubiquitous in a variety of platforms from healthcare to finance to social networks to manufacturing systems~\citep{he2017learning}. 
%For example, a client profile could be characterized by heterogeneous data types including numbers (age/height), labels (gender), text (bio) and images (profile picture). 
By the nature of the observation mechanism of the data, partial observability is often associated with heterogeneity. For example, a bank client is more likely to conceal the income amount than the attributes like gender or age; a doctor can never perform all medical tests for a patient, but choose to collect test results based on particular symptoms and expertise.
In this work we present a deep latent variable model for representation learning from {partially-observed heterogeneous} data.

Deep generative models have been shown to be effective in a variety of homogeneous data learning~\citep{bando2018statistical, gulrajani2017improved, mescheder2017adversarial, yang2017improved}. However, learning these models from heterogeneous data introduces new challenges. In particular, our definition of heterogeneity spans a wide range of forms from data type to data source/modality. For example, we may have categorical or numerical data from different distributions, or mixed modalities representing images, text, and audio. The main challenge is how to align and integrate heterogeneous data to model the joint distribution. Our proposed latent variable model handles this effectively by selecting appropriate proposal distributions, and performing the integration in a latent space instead of the input space.

Learning from partially-observed data is another challenge in deep generative models. 
Naive solutions such as ignoring or zero-imputing missing data will likely degrade performance by introducing sparsity bias~\citep{abs-1906-00150}. 
Having a model designed to learn from incomplete data not only increases the application spectrum of deep learning algorithms but also benefits down-stream tasks such as data imputation, which remains an open and challenging area of research.  

Some prior work requires fully-observed data for training~\citep{suzuki2016joint,ivanov2018variational}, or makes the assumption that data are missing completely at random (MCAR)~\citep{yoon2018gain,li2019misgan}, which assumes \textit{missingness} (the manner in which data are missing) occurs independently from the data. Our method relaxes these assumptions by learning the joint distribution of data and missingness patterns (or \textit{mask}).

In this work, we propose the variational selective autoencoder (VSAE), a general and flexible model for representation learning from partially-observed heterogeneous data. The proposed deep latent variable model is capable of capturing hidden dependencies within partially-observed heterogeneous data by performing selection and integration in the latent representation. 
It learns the joint distribution of data/mask without strong assumptions about missingness mechanism, resulting in applications for data generation and imputation. In particular, it can be trained effectively with a single objective to impute missing data from any combination of observed data.
Extensive evaluation on challenging low-dimensional and high-dimensional heterogeneous data shows improvement over state-of-the-art models.
The contributions are summarized as follows:
\vspace{-0.06in}
\begin{itemize}[leftmargin=*]
  \item A novel selective proposal distribution efficiently learns representations from partially-observed heterogeneous data.
  %\vspace{-0.06in}
  \item The proposed method models the joint distribution of the data and the imputation mask, resulting in a unified model for various tasks including generation and imputation.
  %\vspace{-0.06in}
  \item VSAE does not make restrictive assumptions on the missingness mechanism, expanding the scope of scenarios in which data imputation can be effectively learned.
\end{itemize}

\section{Related Work}
\label{sec:related_work}
\paragraph{Learning from Heterogeneous Data.}

Most existing approaches modeling statistical dependencies in unstructured heterogeneous data focus on obtaining alignments and subsequently modeling relationships between different domains \citep{kim2017learning,zhu2017unpaired,castrejon2016learning}. However, there has been little progress in learning the joint distribution of the full data comprising different domains. 
Other methods~\citep{misra2016cross,zhang2018learning} handle heterogeneity in labels or datasets in weakly-supervised learning settings. 
MVAE~\citep{Wu2018MGM} uses
a product-of-experts inference network to solve the inference problem in multi-model setting. 
Our work focuses on modeling all types of heterogeneity and demonstrates effectiveness in data generation and imputation applications.

\paragraph{Learning from Partially-Observed Data.}

Classical methods dealing with missing data such as MICE~\citep{buuren2010mice} and MissForest~\citep{10.1093/bioinformatics/btr597} typically learn discriminative models to impute missing features. Advanced by deep neural networks, several models have also been developed to address data imputation based on autoencoders ~\citep{gondara2017multiple,vincent2008extracting}, generative adversarial networks (GANs)~\citep{ li2019misgan,yoon2018gain}, and autoregressive models \citep{bachman2015data}.
In this work, we focus on improving deep latent variable models to efficiently learn from partially-observed heterogeneous data.
Deep latent variable models (DLVMs) are generative models that can map complex raw input to a flexible latent representation and have recently gained attention on handling partially-observed data due to the flexibility of generative modeling and representation learning. 
To tackle the intractable posterior of DLVMs, variational autoencoder (VAE) first uses deep neural networks to approximate the posterior and maximizes a variational evidence lower bound (ELBO). 
Based on VAE, prior work~\citep{ivanov2018variational, MaTPHNZ19, mattei2019miwae, nazabal2018handling, collier2020vaes} attempted to improve DLVMs under strong missingness mechanism assumptions.
VAEAC~\citep{ivanov2018variational} imputed attributes conditional on observed ones by learning from fully-observed data; 
Partial VAE~\citep{MaTPHNZ19} encoded observed data with a permutation invariant set function;
MIWAE~\citep{mattei2019miwae} introduced a tighter bound using importance sampling under a relaxed Missing At Random (MAR, refer to Sec.~\ref{sec:pre_imp-process}) assumption;
HI-VAE~\citep{nazabal2018handling} factored decoders on low-dimensional heterogeneous data under MCAR. 
These methods jointly map the sources (a.k.a.\ \textit{attribute}) of heterogeneous data into a holistic latent space, which adds unavoidable noise to the latent space as the distribution of one attribute can be far from the others.

\section{Background}
\label{sec:method}

\textbf{Problem Statement.}
We represent any \textit{heterogeneous} data point as a set of random variables $\x=[\x_{1},\x_{2}...,\x_{M}]$ representing different \textit{attributes} collected from multiple sources. The type and size of each attribute $\x_i$ can vary. It can be either high-dimensional (e.g.\ multimedia data) or low-dimensional (e.g.\ tabular data). 
We define an $M$-dimensional binary \textit{mask} variable $\mathbf{m} \in \mathbf\{0,1\}^M $ to represent the missingness: for the $i$-th attribute, $m_i=1$ if it is observed and $0$ otherwise.
Thus we can induce \textit{observed attributes} by the set $\mathbb{O} = \{i|m_i=1\}$ and \textit{unobserved attributes} by the complementary set $\mathbb{U}=\{i|m_i=0\}$. 
Accordingly, we denote the collective representation of the observed attribute with $\x_\mathbf{o} = [\x_i|m_i=1]$ and unobserved attributes with $\x_\mathbf{u} = [\x_i|m_i=0]$. 
In general, every instance has a different set of $\x_\oo$ as well as $\x_\uu$, determined by the mask variable $\mm$.
% Our goal is to learn the joint distribution of all attributes and mask together from incomplete training data.

\textbf{Missingness Mechanism. }
\label{sec:pre_imp-process}
The generative process of incomplete data can be modeled by the joint distribution 
 $ p(\x_\mathbf{o}, \x_\mathbf{u}, \mathbf{m})$. %and learned in a marginalized maximum likelihood setting $
%     \max_{\boldsymbol{\lambda}}
%   % \int p(\x_\mathbf{o}, \x_\mathbf{u}, \mathbf{m}) d\x_\mathbf{u}= 
%     \int p_{\boldsymbol{\lambda}}(\x_\mathbf{o}, \x_\mathbf{u}, \mathbf{m}) d\x_\mathbf{u} = \max_{\boldsymbol{\lambda}} p_{\boldsymbol{\lambda}}({\x_\mathbf{o}, \mathbf{m}}).$
\cite{little2019statistical} categorize the missingness mechanism into three types based on the dependence between the data and mask as follows,

\textbf{\textit{Missing Completely At Random (MCAR).}} Missingness is completely independent of data, 
\vspace{-1mm}
\begin{align}\label{eq:mcar}
    p(\x_\mathbf{o}, \x_\mathbf{u}, \mathbf{m}) = p(\x_\mathbf{o}, \x_\mathbf{u}) p(\mathbf{m})
\end{align}
\textbf{\textit{Missing At Random (MAR).}} Missingness depends only on observed attributes, 
\vspace{-1mm}
\begin{align}\label{eq:mar}
    p(\x_\mathbf{o}, \x_\mathbf{u}, \mathbf{m}) = p(\x_\mathbf{o}, \x_\mathbf{u}) p(\mathbf{m}|\x_\mathbf{o})
\end{align}
\textbf{\textit{Not Missing At Random (NMAR).}} Missingness depends on both observed and unobserved attributes,
\vspace{-1mm}
\begin{align}\label{eq:nmar}
    p(\x_\mathbf{o}, \x_\mathbf{u}, \mathbf{m}) = p(\x_\mathbf{o}, \x_\mathbf{u}) p(\mathbf{m}|\x_\mathbf{o}, \x_\mathbf{u}) 
    % \text{ or } 
    % p(\x_\mathbf{o}, \x_\mathbf{u})p(\mathbf{m}|\x_\mathbf{u})
\end{align}
Most prior work on learning from partially-observed data follows the MCAR or MAR assumption since the factorization in Eq.~\eqref{eq:mcar} and Eq.~\eqref{eq:mar} decouples the mask $\mathbf{m}$ from $\x_\mathbf{u}$ in the integral of the likelihood function. A common approach is to define the log-likelihood of an incomplete dataset by marginalizing over the unobserved attributes $\log [\int p( \x_\oo,\x_\uu, \mm) d\x_\uu] = \log p(\x_\oo, \mm)$ and ignoring the underlying missing mechanism. Our goal is to go beyond this simple but restrictive solution and model the joint distribution of the data and the mask.

%a simple but restrictive solution.
\section{Proposed Method}

\label{sec:math-vsae}

\begin{figure*}[t]
\centering
\includegraphics[width=1\textwidth]{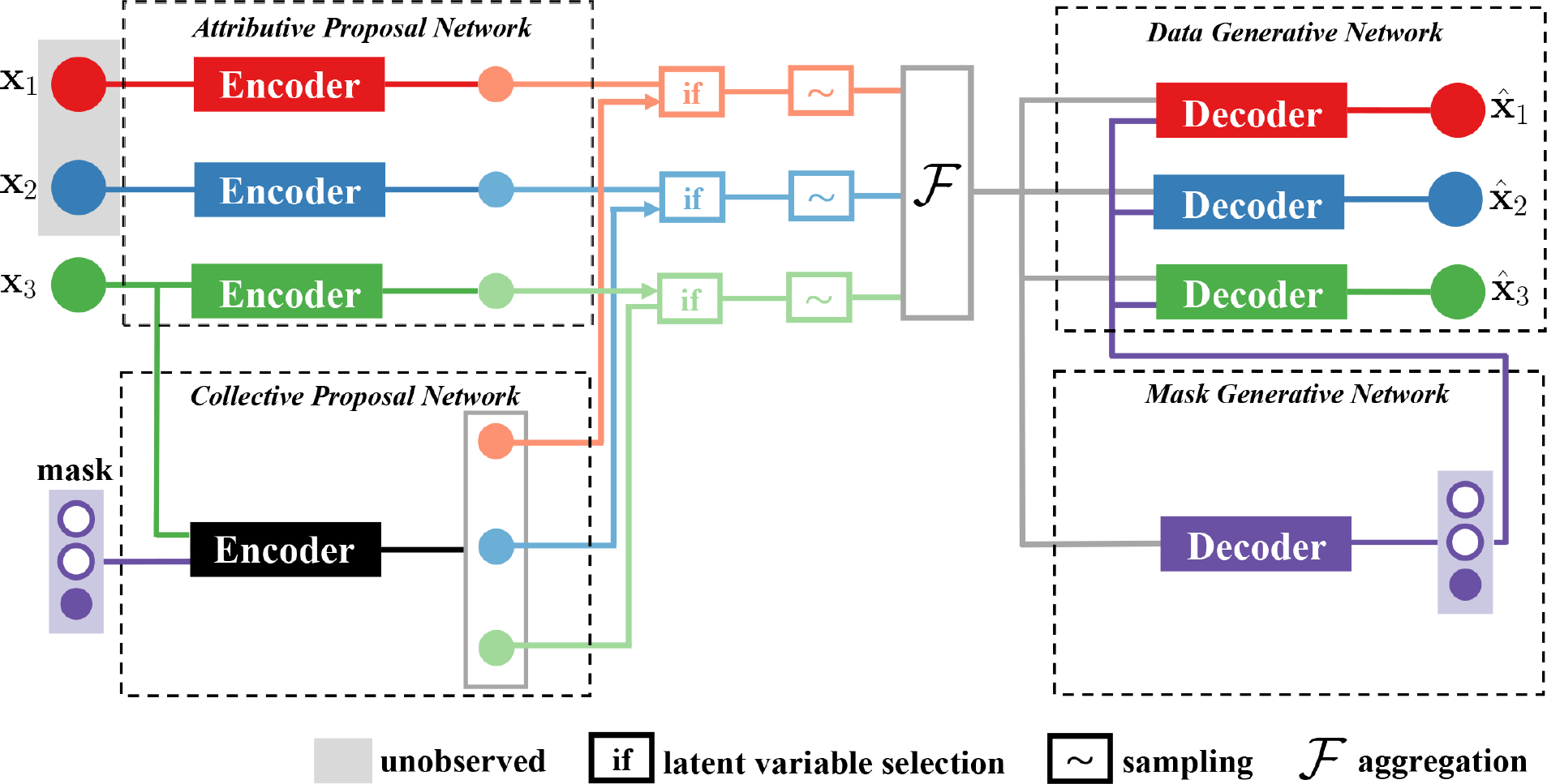}
\caption{\textbf{Model overview.} The input includes data \textit{attributes} (denoted by different colors --- $\x_1,\x_2$ are unobserved; $\x_3$ is observed) and \textit{mask}. The \textbf{\textit{attributive proposal network}} takes the attributes as input, while the \textbf{\textit{collective proposal network}} takes the observed attributes and the mask as input. We use $\operatorname{if}$-condition to denote the selective proposal distribution (Eq.~\ref{eq:selective_proposal}, the arrowed one is selected). The aggregated latent codes are fed to both \textbf{\textit{mask generative network}} and \textbf{\textit{data generative network}}. The output of the mask generative network will be fed to each decoder of the data generative network as extra condition. Standard normal prior is not plotted for simplicity.
All components are trained simultaneously in an end-to-end manner.}
\label{fig_arc}
\end{figure*}

\textcolor{black}
Variational Selective Autoencoder (VSAE) aims to relax the strong missingness assumption and model the joint distribution $p(\x, \mathbf{m})=\int p(\x, \mathbf{m}| \mathbf{z})p(\mathbf{z})d\mathbf{z}$. The latent variables $\mathbf{z}$ are inferred from data instances with the mask to capture the dependence among the attributes and the mask. The data and missingness information embedded in the latent variables can together guide data imputation and generation.
At a high level, as illustrated in Fig.~\ref{fig_arc}, VSAE is formulated with individual encoders for observed attributes and a collective encoder for unobserved ones to construct the selective proposal distribution. The selected latent variables of each attribute are aggregated and decoded to reconstruct the mask and all attributes independently. To tackle the intractability of unobserved attributes $\x_\mathbf{u}$ and latent variables $\mathbf{z}$, we formulate the learning procedure as an external expectation maximization (EM) nested with variational inference (VI). In the following, we provide details on the VSAE model.

\subsection{Model description}
Following the VAE\footnote{Refer to Appendix A for a detailed description of VAE. }~\citep{KingmaW13,pmlr-v32-rezende14}, we construct a proposal distribution $q(\mathbf{z}|\x, \mathbf{m})$ to approximate the intractable true posterior. 
With the inclusion of the novel selective proposal distribution, we expand the parameters of inference networks to $\{\boldsymbol{\phi},\boldsymbol{\psi}\}$, where  $\boldsymbol{\phi}$ and $\boldsymbol{\psi}$ represent encoder parameter for observed and unobserved attributes. Following the same fashion, the parameters of generative networks are expanded to $\{\boldsymbol{\theta},\boldsymbol{\epsilon}\}$, with $\boldsymbol{\theta}$ denoting the decoder parameter for the data, and $\boldsymbol{\epsilon}$ for the mask.
The variational evidence lower bound of $\log p(\x, \mathbf{m})$ can thus be derived as
%\vspace{-2mm}
\begin{align}\label{eq:elbo_init}
\begin{aligned}
&\mathcal{L}_{\boldsymbol{\phi}, \boldsymbol{\psi}, \boldsymbol{\theta}, \boldsymbol{\epsilon}}(\x,\mathbf{m})=%\nonumber
\underbrace{\mathbb{E}_{\mathbf{z}\sim q_{\boldsymbol{\phi}, \boldsymbol{\psi}}(\mathbf{z}|\x, \mathbf{m})}[{\log p_{\boldsymbol{\theta}, \boldsymbol{\epsilon}}(\x,\mathbf{m}|\mathbf{z})}]}_\text{cond. LL}\\&- 
\underbrace{{D}_\text{KL}
{(q_{\boldsymbol{\phi}, \boldsymbol{\psi}}(\mathbf{z} | \x, \mathbf{m}) || p(\mathbf{z})),}}_\text{KL Regularizer}
\end{aligned}
\end{align}
%where the KL divergence 
% ${D}_\text{KL}{[q_{\boldsymbol{\phi}, \boldsymbol{\psi}}(\mathbf{z} | \x, \mathbf{m}) || p(\mathbf{z})]} =
% \mathbb{E}_{\mathbf{z}\sim q_{\boldsymbol{\phi}, \boldsymbol{\psi}}(\mathbf{z}|\x, \mathbf{m})}[\log q_{\boldsymbol{\phi}, \boldsymbol{\psi}}(\mathbf{z} | \x, \mathbf{m}) -
% \log p(\mathbf{z})]$ 
%acts as a regularizer to push proposal distribution $q_{\boldsymbol{\phi}, \boldsymbol{\psi}}(\mathbf{z} | \x, \mathbf{m})$ close to prior $p(\mathbf{z})$. 

\textbf{Factorized Latent Space.} We assume the latent space can be factorized w.r.t.\ $M$ attributes,% that $\mathbf{z}=[\mathbf{z}_1, \mathbf{z}_2, ..., \mathbf{z}_M]$, 
%\vspace{-6pt}
\begin{align}\label{eq:factorization}
     p(\mathbf{z})&=\prod_{i=1}^M p(\mathbf{z}_i), \quad
    q(\mathbf{z}|\x, \mathbf{m})=\prod_{i=1}^M q(\mathbf{z}_i|\x, \mathbf{m})
\end{align}
Priors $p(\mathbf{z}_i)$ are standard Gaussians and proposal distributions $q(\mathbf{z}_i|\x, \mathbf{m})$ are Gaussians with inferred means and diagonal covariances. This factorization separates the encoding of each attribute and efficiently yields a distribution for latent variables by assuming the latent variables are conditionally independent given the data and mask. Hence, it provides a mechanism to decouple the heterogeneity in the raw data space while integrating them efficiently in the latent space. 

\textbf{Selective Proposal Distribution.} The standard proposal distribution of VAEs, inferred from fully-observed data, is not applicable for partially-observed input. To circumvent this, we introduce our \textit{selective proposal distribution} for each latent variable: 
%\vspace{-2pt}
\begin{equation}\label{eq:selective_proposal}
  q_{\boldsymbol{\phi},\boldsymbol{\psi}}(\mathbf{z}_i|\x, \mathbf{m}) =
    \begin{cases}
    q_{\boldsymbol{\phi}}(\mathbf{z}_i|\x_i) & \text{if $m_i=1$}\\
    q_{\boldsymbol{\psi}}(\mathbf{z}_i|\x_\mathbf{o},\mathbf{m}) & \text{if $m_i=0$}\\
    \end{cases}       
\end{equation}
This conditional selection of proposal distribution is determined by the mask variable. Accordingly, we subdivide the inference network into two types:

\textbf{\textit{Attributive Proposal Network.}}
 $q_{\boldsymbol{\phi}}(\mathbf{z}_i | \x_i)$, inferred merely from the individual observed attribute and selected for an observed attribute;
 
\textbf{\textit{Collective Proposal Network.}} $q_{\boldsymbol{\psi}}(\mathbf{z}_i | \x_\mathbf{o}, \mathbf{m})$, collecting all observed values and the mask to produce the proposal distribution and selected for an unobserved attribute. 
This formulation aids VAE encoders by explicitly focusing on the relevant inputs and ignoring the less informative ones.

\textbf{Latent Variable Aggregation.}
We sample the latent variables for all attributes using Eq.~\eqref{eq:selective_proposal}.
Next, to capture the dependencies between observed attributes, unobserved attributes and mask, an aggregation function $\mathcal{F}(\cdot)$ is performed before the decoders.
% , so that
% $p_{\boldsymbol{\epsilon}}(\mathbf{m}|\mathbf{z}) = p_{\boldsymbol{\epsilon}}(\mathbf{m}|\mathcal{F}(\mathbf{z}))$, $p_{\boldsymbol{\theta}}(\x_i|\mathbf{z},\mathbf{m}) = p_{\boldsymbol{\theta}}(\x_i|\mathcal{F}(\mathbf{z}),\mathbf{m}))$.
We use \textit{concatenation} as $\mathcal{F}(\cdot)$, though it can be any aggregation function in general.
The conventional VAEs, however, often aggregate the attributes naively in the raw data space. Consequently, the heterogeneity and partially-observed nature will restrain those models from learning informative representations.

\textbf{Data \& Mask Generative Networks.}
By applying the chain rule, the conditional log-likelihood $\log p_{\boldsymbol{\theta}, \boldsymbol{\epsilon}}(\x,\mathbf{m}|\mathbf{z})$ in Eq.~\eqref{eq:elbo_init} is decomposed as {mask conditional log-likelihood} $\log p_{\boldsymbol{\epsilon}}(\mathbf{m}|\mathbf{z})$ and {data conditional log-likelihood} $\log p_{\boldsymbol{\theta}}(\x|\mathbf{m}, \mathbf{z})$. The mask and data are reconstructed from shared $\zz$ through the \textbf{\textit{mask generative network}} and \textbf{\textit{data generative network}} shown in Fig.~\ref{fig_arc}. 
Further, the data conditional log-likelihood factorizes over the attributes assuming the reconstructions are conditionally independent given $\mathbf{m}$ and $\mathbf{z}$:
%\vspace{-5pt}
\begin{align}\label{eq:generative_independence}
\medmath{\log p_{\boldsymbol{\theta}}(\x|\mathbf{m}, \mathbf{z}) %\nonumber\\
=\underbrace{\sum_{i \in \mathbb{O}} \log p_{\boldsymbol{\theta}}(\x_i|\mathbf{m}, \mathbf{z})}_\text{Observed} + \underbrace{\sum_{j \in \mathbb{U}} \log p_{\boldsymbol{\theta}}(\x_j|\mathbf{m}, \mathbf{z})}_\text{Unobserved}
}
\end{align} 

\textcolor{black}{\textbf{Expectation Maximization.}} The ELBO (Eq.~\eqref{eq:elbo_init}) is hard to maximize since $\x_\mathbf{u}$ is unobserved. \textcolor{black}{We can use expectation maximization (EM) algorithm~\citep{Dempster77maximumlikelihood} to handle its intractability (refer to Appendix B for full derivation). EM alternates between inferring the unobserved data given the parameters ($E$ step) and optimizing the parameters given the “filled in” data ($M$ step).}

\textbullet {\textbf{ $E$ step:}} By taking an expectation over $\x_\mathbf{u}$,
%\vspace{-6pt}
\begin{align}\label{eq:expected_obj}
\mathcal{L}'_{\boldsymbol{\phi}, \boldsymbol{\psi}, \boldsymbol{\boldsymbol{\theta}}, \boldsymbol{\boldsymbol{\epsilon}}}(\x_\mathbf{o},\mathbf{m})= \mathbb{E}_{\x_\mathbf{u}}[\mathcal{L}_{\boldsymbol{\phi}, \boldsymbol{\psi}, \boldsymbol{\boldsymbol{\theta}}, \boldsymbol{\boldsymbol{\epsilon}}}(\x_\mathbf{o},\x_\mathbf{u},\mathbf{m})]
\end{align}
%We can plug Eq.~\eqref{eq:elbo_init} \textasciitilde \eqref{eq:generative_independence} into Eq.~(\ref{eq:expected_obj}).  
We now expand Eq.~\eqref{eq:expected_obj} into conditional log-likelihood and KL divergence terms. 
Since only the unobserved attributes conditional log-likelihood depends on $\x_\mathbf{u}$, with $ \mathbf{z}_i\sim q_{\boldsymbol{\phi}, \boldsymbol{\psi}}(\mathbf{z}_i|\x, \mathbf{m})$ given by Eq.~(\ref{eq:selective_proposal}), we obtain
\begin{align}\label{eq:elbo_final}
&\mathcal{L}'_{\boldsymbol{\phi}, \boldsymbol{\psi}, \boldsymbol{\boldsymbol{\theta}}, \boldsymbol{\boldsymbol{\epsilon}}}(\x_\mathbf{o},\mathbf{m})= 
 \underbrace{\mathbb{E}_{\mathbf{z}}\big[
    \sum_{j\in \mathbb{U}} \mathbb{E}_{\x_j}[\log p_{\boldsymbol{\theta}}(\x_j|\mathbf{m},\mathbf{z})]}_\text{Unobserved attributes cond. LL} \big]\nonumber
    +\\&
   \underbrace{ \mathbb{E}_{\mathbf{z}}
    \big[\sum_{i\in \mathbb{O}} \log p_{\boldsymbol{\theta}}(\x_i|\mathbf{m},\mathbf{z})\big]}_\text{Observed attributes cond. LL}
    +
    \underbrace{\mathbb{E}_{\mathbf{z}}[
    \log p_{\boldsymbol{\epsilon}}(\mathbf{m}|\mathbf{z})]}_\text{Mask cond. LL}\nonumber -\\
    &
    \underbrace{\sum_{i=1}^{M} \mathbb{E}_{\mathbf{z}_i}[\log { q_{\boldsymbol{\phi}, \boldsymbol{\psi}}(\mathbf{z}_i|\x, \mathbf{m})}-\log { p(\mathbf{z}_i)}]}_\text{KL Regularizer} 
\end{align}
% \begin{align}\label{eq:elbo_final}
% &\mathcal{L}'_{\boldsymbol{\phi}, \boldsymbol{\psi}, \boldsymbol{\boldsymbol{\theta}}, \boldsymbol{\boldsymbol{\epsilon}}}(\x_\mathbf{o},\mathbf{m})= 
% \underbrace{\mathbb{E}_{\mathbf{z}}[
%     \log p_{\boldsymbol{\epsilon}}(\mathbf{m}|\mathbf{z})]}_\text{Mask cond. LL}\nonumber
%     \\&+
%   \underbrace{ \mathbb{E}_{\mathbf{z}}
%     \big[\sum_{i\in \mathbb{O}} \log p_{\boldsymbol{\theta}}(\x_i|\mathbf{m},\mathbf{z})\big]}_\text{Observed attributes cond. LL}
%     + \underbrace{\mathbb{E}_{\mathbf{z}}\big[
%     \sum_{j\in \mathbb{U}} \mathbb{E}_{\x_j}[\log p_{\boldsymbol{\theta}}(\x_j|\mathbf{m},\mathbf{z})]}_\text{Unobserved attributes cond. LL} \big]\nonumber \\
%     &-
%     \underbrace{\sum_{i=1}^{M} \mathbb{E}_{\mathbf{z}_i}[\log { q_{\boldsymbol{\phi}, \boldsymbol{\psi}}(\mathbf{z}_i|\x, \mathbf{m})}-\log { p(\mathbf{z}_i)}]}_\text{KL Regularizer} 
% \end{align}
In Eq.~(\ref{eq:elbo_final}), direct calculation of unobserved attributes conditional log-likelihood is intractable. Instead, during training given the learned generative model $p(\x_\oo,\x_\uu,\mm,\zz)$ with VI, we can generate conditional distribution of unobserved attribute $\x_j$ from:
% \vspace{-10pt}
\begin{align}\label{eq:syn_unob}
\medmath{
    \hat{p}({\x}_j|\x_\oo, \mm)= \int
     \Phi(\mathbf{z})
    p_{\boldsymbol{\theta}^*}(\x_j|\mathbf{m}, \mathbf{z})
    d\mathbf{z}
    }
\end{align} 
 where $\Phi(\zz):=
 \prod_{i \in \mathbb{O}}  q_{\boldsymbol{\phi}^*}(\zz_i|\x_i)
 \prod_{j \in \mathbb{U}}  p(\zz_i)$ and $*$ denotes the parameters learned up to the current iteration. It enables us to estimate each unobserved terms with 
$\mathbb{E}_{\x_j\sim\hat{p}_(\x_
j|\x_\oo, \mm)}[\log p_{\boldsymbol{\theta}}(\x_j|\mathbf{m},\mathbf{z})]$.

\textbullet\textbf{ $M$ step:} We can therefore maximize the final objective function (Eq.~(\ref{eq:elbo_final})) with $\x_\uu$ sampled from Eq.~(\ref{eq:syn_unob}). 
Conventional methods of learning from partially-observed data simply marginalizes and  maximizes $\log \int p(\x_\oo,\x_\uu,\mm)d\x_\uu = \log p(\x_\oo, \mm)$ by ignoring $\x_\uu$. Our procedure can be viewed as maximizing the expected lower bound $\mathbb{E}_{\x_\mathbf{u}}[\mathcal{L}_{\boldsymbol{\phi}, \boldsymbol{\psi}, \boldsymbol{\boldsymbol{\theta}}, \boldsymbol{\boldsymbol{\epsilon}}}(\x_\mathbf{o},\x_\mathbf{u},\mathbf{m})]$
to approximately maximize
 $\mathbb{E}_{\x_\uu} [\log p(\x_\oo, \x_\uu,\mathbf{m})]$.
 
Empirically, given a partially-observed batch, our two-stage scheme is: 
(1) decode the latent codes drawn from the prior $p(\zz_i)$ of unobserved attributes and the attributive proposal distribution $q_{\boldsymbol{\phi}^*}(\mathbf{z}_i|\x_i)$ of observed attributes to generate $\x_\uu$ by the decoders learned so far; 
(2) re-input the same incomplete batch to calculate all observed terms and estimate the unobserved term with Eq.~(\ref{eq:syn_unob}), then optimize over all parameters. 
 We generate 100 samples of $\x_\uu$ to take expectation and experiments show that it gives an effective estimation to the full expectation. 
% The prior network can encourage the model itself to find the most likely unobserved attributes of this instance by taking the expectation. 

Another possible alternative in $E$ step is to infer the latent variables of unobserved attributes from the collective proposal distribution $q_{\boldsymbol{\psi^*}}(\mathbf{z}|\x_\oo, \mathbf{m})$ but empirically we find the prior performs better. The possible explanation lies in two-fold: (i) Approximating the true posterior $p(\mathbf{z}|\mathbf{x}_\mathbf{u}, \mathbf{x}_\mathbf{o}, \mathbf{m})$ is challenging without $\mathbf{x}_\mathbf{u}$. At early stages of training, the parametric proposal distribution is not learned well to match the posterior. Hence, the parameter-free prior is a more stable choice for the noisy unobserved attributes;
% \textbf{(ii)} We agree that the prior is uninformed, but it is only used for the latent variables of $\mathbf{x}_\mathbf{u}$. The aggregated latent variables are still conditioned on $\mathbf{x}_\mathbf{o}$ from the attributive proposals;
%  The prior is static during the training. 
%  In the early stage, 
%  The optimal posterior is $p(z|x_u, x_o, m)$, which we approximate with our proposal distribution.
%  \textbf{(ii)} Approximating the true posterior $p(\mathbf{z}|\mathbf{x}_\mathbf{u}, \mathbf{x}_\mathbf{o}, \mathbf{m})$ is challenging without $\mathbf{x}_\mathbf{u}$. Early in learning, our learned proposal distribution does not match the posterior well;
%  The poorly learned encoders may further negatively affect the decoders; 
%  For$\mathbf{x}_\mathbf{u}$, sampling from the fixed prior is possibly regarded as adding some random noise to the decoders, but sampling from immature posterior of unobserved attributes may introduce some bias to falsely guide the model to learn a meaningful joint distribution.  
 {(ii)} If sampling from the prior for unobserved attributes, the aggregated latent variables are still conditioned on $\mathbf{x}_\mathbf{o}$ from the attributive proposal distribution. Compared to collective proposal distribution $q_{\boldsymbol{\psi^*}}(\mathbf{z}|\x_\oo, \mathbf{m})$ for $\mathbf{x}_\mathbf{u}$, sampling from the prior does not depend on $\mathbf{m}$. Since both data and mask generative processes have not yet been accurately captured, it can encourage focus on generating $\mathbf{x}_\mathbf{u}$.

\subsection{Model applications}\label{subsec:model_app}
Unlike conventional data imputation models, the generative model $p(\x,\mathbf{m},\mathbf{z})$ is learned. Therefore, VSAE constructs a unified framework for data imputation, data generation and mask generation. 

\textbf{{Data Imputation. }}The aim is to impute the missing data given the observed data, thus it can be viewed as conditional generation of the unobserved attributes. 
 This can be performed by sampling the latent codes for all attributes using $q_{\boldsymbol{\phi}, \boldsymbol{\psi}}(\mathbf{z}_i|\x, \mathbf{m})$ in Eq.~(\ref{eq:selective_proposal}). Next, the aggregated latent codes and mask are given to the decoders of the unobserved attributes for generation.
This process can be described as 
$ p(\x_{\mathbf{u}} | \x_{\mathbf{o}}, \mathbf{m}) \approx
    \int p_{\boldsymbol{\theta}}(\x_{\mathbf{u}}|\mathbf{m},\mathbf{z})
    q_{\boldsymbol{\phi},\boldsymbol{\psi}}(\mathbf{z}|\x_{\mathbf{o}},\mathbf{m}) d\mathbf{z}$.

\textbf{{Data \& Mask Generation.}}
% We can obtain by chain rule that
% $
%     p_{\boldsymbol{\theta},\boldsymbol{\epsilon}}(\x,\mathbf{m},\mathbf{z})=
%     p_{\boldsymbol{\theta}}(\x|\mathbf{m},\mathbf{z})
%     p_{\boldsymbol{\epsilon}}(\mathbf{m}|\mathbf{z})
%     p(\mathbf{z})
% $, therefore
% \begin{align}\label{eq:generation_eq}
%     &p_{\boldsymbol{\theta},\boldsymbol{\epsilon}}(\x,\mathbf{m})=
%     \int p_{\boldsymbol{\theta}}(\x|\mathbf{m},\mathbf{z})
%     p_{\boldsymbol{\epsilon}}(\mathbf{m}|\mathbf{z})
%     p(\mathbf{z}) d\mathbf{z}
% \end{align}
% where $\boldsymbol{\theta}$ and $\boldsymbol{\epsilon}$ are specified as parameters of data generative network and mask generative network. 
Given random samples from a standard Gaussian prior $p(\mathbf{z})$, we can generate masks using the mask generative network. Next, the sampled latent codes and the generated mask are given to the data generative network to generate data.

Modeling the mask distribution in our framework not only helps to inform the data generation of the relevant missingness mechanism but also enables to sample the mask itself. MisGAN~\citep{li2019misgan} also employs a mask generation process. However, MisGAN generates the mask assuming the mask is MCAR, whereas VSAE generates the mask assuming the mask is NMAR.  Mask generation may have applications like synthesizing incomplete data or obtaining the potential mask if the real mask is not available (e.g.\ if data is corrupted rather than only missing). 

\subsection{Scalability}
The complexity grows linearly w.r.t the number of attributes $M$ and their dimensions. In practice, heterogeneous data are often either \textit{tabular} (large $M$ but low-dimensional) or \textit{multi-modal} (high-dimensional but small $M$), thus practically tractable. For large $M$ with high-dimensional attributes (e.g. URLs), one may use embedding layers to densify these categorical attributes to low-dimensional vectors, or manually group subset of the attributes to constrain the $M$ as a feasible constant.
\begin{figure}[t]%{r}%{0.5\textwidth}
\centering
\includegraphics[width=0.98\textwidth]{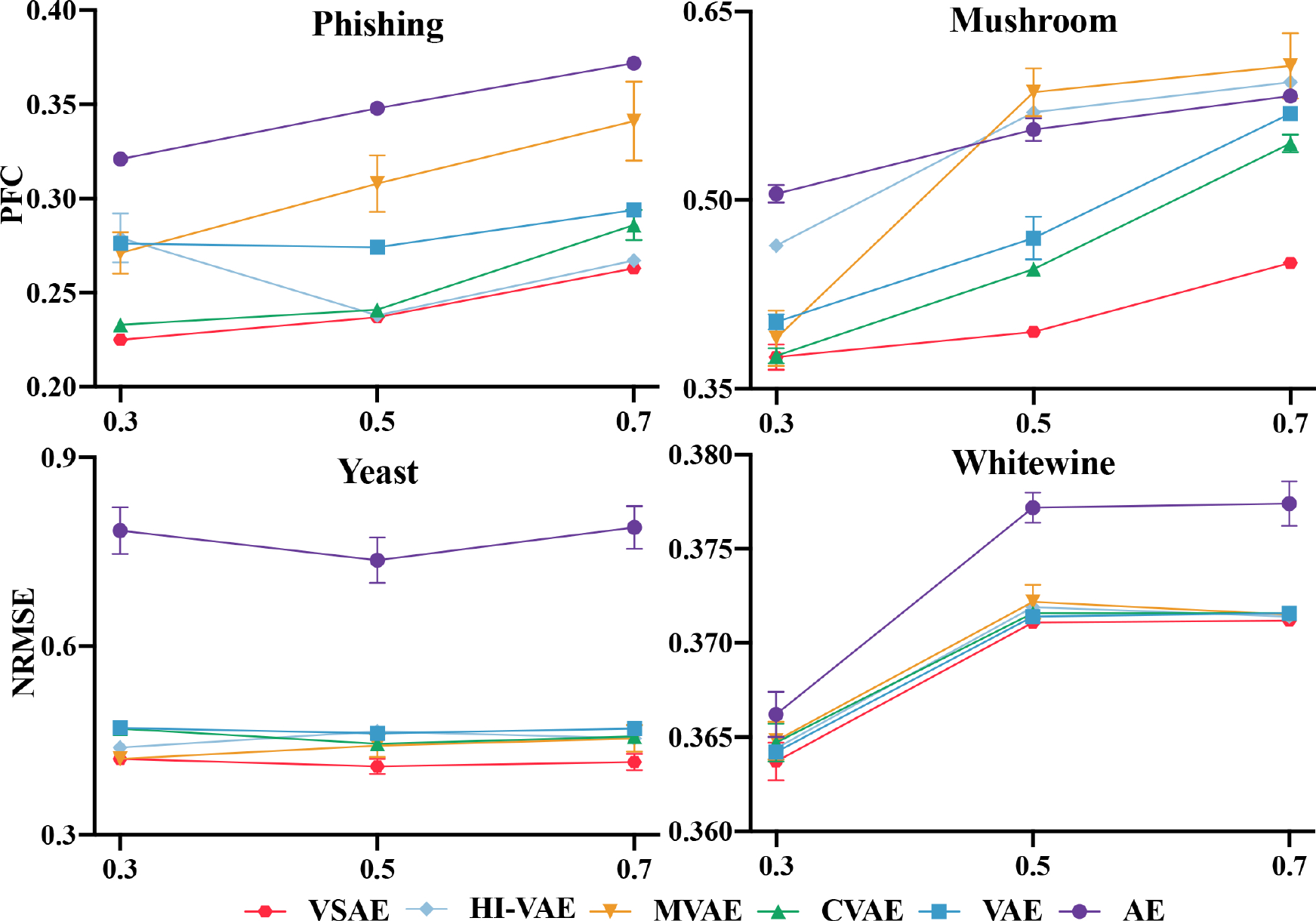}
\caption{\textbf{Data Imputation on UCI datasets.} X-axis is missing ratio. Categorical (top)/numerical (bottom) datasets are evaluated by \textbf{PFC}/\textbf{NRMSE} with mean/std over 3 independent trials. Lower is better.}\label{fig:uci_ratios}
\end{figure}

\section{Experiments}
\label{experiment}
%We consider data imputation, data generation and mask generation on low-dimensional \textit{tabular data} and high-dimensional \textit{multi-modal data} under various missing mechanisms.
%In tabular datasets,  heterogeneous data consists of numerical and categorical attributes. In multi-modal datasets, the input variables are high-dimensional representations of image, text, or audio from different distributions. 

%in comparison with state-of-the-art models. 

\subsection{Tabular Data}\label{sec::tabular_data}

\begin{table*}[t]
\centering
\caption{\textbf{MCAR Data Imputation on UCI datasets.} We consider categorical, numerical and mixed tabular datasets in the missing ratio of 0.5. Categorical and numerical attributes are evaluated by \textbf{PFC} and \textbf{NRMSE} respectively, lower is better for both. We show mean and standard deviation over 3 independent trials.  $\Delta<0.0005$.}
\label{table:uci_mcar}
\scalebox{0.89}{\begin{tabular}{lcccccccc}
\toprule
%\cmidrule{2-3}\cmidrule{5-6}
 & \textbf{Phishing}&\textbf{Mushroom} & &\textbf{Yeast}&\textbf{Whitewine} &&  \multicolumn{2}{c}{\textbf{Heart (mixed)}} \\
\cmidrule{1-3}\cmidrule{5-6}\cmidrule{8-9}
 {Attribute type} &categorical&categorical  &&numerical&numerical &&categorical&numerical \\
\midrule
AE 
& $0.348\pm0.004$ & $0.556\pm0.009$ & 
& $0.737\pm0.035$ & $0.3772\pm\Delta$&
& $0.514\pm0.026$ & $0.650\pm0.006$ \\
VAE
& $0.274\pm\Delta$ & $0.470\pm0.016$  &
& $0.461\pm\Delta$ & $0.3724\pm\Delta$ &
& $\bf{0.485\pm0.016}$ & $0.630\pm 0.025$\\
CVAE w/ mask 
& $0.241\pm\Delta$ & $0.445\pm\Delta$ & 
& $0.445\pm\Delta$ & $0.3726\pm\Delta$& 
& $0.516\pm 0.015$& $0.605\pm0.005$\\
MVAE 
& $0.308\pm0.016$ & $0.586\pm0.017$ &
& $0.442\pm0.016$ & $0.3732\pm\Delta$&
& $0.517\pm0.010$ & $0.636\pm0.009$\\ 
HI-VAE 
& $0.238\pm\Delta$ & $0.470\pm0.008$ & 
& $0.429\pm0.005$ & $\bf{0.3720\pm\Delta}$&
& $\bf{0.481\pm0.012}$ & $0.603\pm0.008$\\

VSAE (ours)
& $\bf{0.237\pm\Delta}$ & $\bf{0.416\pm0.009}$ & 
& $\bf{0.419\pm0.008}$  & $\bf{0.3719\pm\Delta}$ &
 & $\bf{0.482\pm0.014}$& $\bf{0.579\pm 0.015}$\\
\bottomrule
\end{tabular}}
\end{table*}

Tabular data are ordered arrangements of rows and columns. Each row is a sample with multiple attributes (typically low-dimensional) and each column is a single attribute collected heterogeneously. Due to communication or privacy issues, those attributes of data samples often are partially-observed. 
For this scenario, we choose \textbf{{UCI repository}} which contains various tabular datasets of numerical or categorical attributes. 
In all experiments, min-max normalization is applied to pre-process the numerical data and the unobserved dimensions are replaced by standard normal noise. 
We split the training and test set with size ratio 4:1 and use 20$\%$ of training data as a validation set to choose the best model. Mean-squared error, cross-entropy and binary cross-entropy are used as reconstruction loss for numerical, categorical and mask variables, respectively.

\paragraph{Data Imputation.}

We first consider a data imputation experiment---imputing unobserved attributes given observed attributes and mask. VSAE can be used in this case as in Sec.~\ref{subsec:model_app}. We report the standard measures: \textbf{{NRMSE}} (RMSE normalized by the standard deviation of ground truth features, averaged over all features) and \textbf{{PFC}} (proportion of falsely classified attributes of each feature, averaged over all features) for numerical and categorical attributes. The evaluation is under various missing mechanisms by synthesizing masks following different rules.

\textbf{\textit{MCAR masking.}} We randomly sample from independent Bernoulli distributions with predefined missing ratios to mimic MCAR missing mechanism. 
VSAE is compared with the deterministic autoencoder (AE), VAE, conditional VAE (CVAE)~\citep{NIPS2015_5775} conditioned on the mask, multi-modal MVAE~\citep{Wu2018MGM} and HI-VAE~\citep{nazabal2018handling}. 
We use publicly released codes of MVAE/HI-VAE, and implement other baselines with the same backbone and at least as many parameters as VSAE.
%\footnote{\textcolor{black}{We also evaluate baselines with their original architectures/hyperparameters but observe no improvement (refer to Appendix E).}} 

%Additional information about experimental details can be found in Appendix B.%~\ref{sec:implementat-details}.

% \begin{wrapfigure}[h]
% \centering
% \includegraphics[width=1\textwidth]{figures/uci_error_bar.pdf}
% \caption{\textbf{Data Imputations on UCI datasets.} Missing ratios (x-axis) are 0.3, 0.5, 0.7. Categorical (top row) and numerical (bottom row) datasets are evaluated by \textbf{PFC} and \textbf{NRMSE} respectively. We show mean and standard deviation over 3 independent runs. For both lower is better.}\label{fig:uci_ratios}
% \end{wrapfigure}

\begin{table}[t]%{r}%{4cm}
\caption{\textbf{Non-MCAR Data Imputation.} We show mean and standard deviation of \textbf{NRMSE} over 3 independent trials, lower is better. $\Delta<0.0005$. } \vspace{-5pt}\label{table:uci_nonmcar}
\scalebox{0.83}{\begin{tabular}{c c  c c} 
 \toprule
&\textbf{Method}  &  MAR & NMAR \\ %[0.5ex] 
\midrule
  \multirow{2}{*}{\textbf{Yeast}} 
  & MIWAE & $\bf{0.475\pm0.005}$ & $0.456 \pm 0.036$\\
  & VSAE (ours)  & $\bf{0.472\pm0.006}$ & $\bf{0.425\pm0.007}$\\
 %NMAR 
 \midrule
 \multirow{2}{*}{\textbf{Whitewine}} 
 & MIWAE & $0.3834\pm\Delta$ & $0.3723 \pm \Delta$\\
 & VSAE (ours)  &$\bf{0.3825\pm\Delta}$ & $\bf{0.3717\pm\Delta}$\\
%   \multirow{2}{*}{\textbf{Whitewine}} 
%  & MIWAE & $0.493\pm\Delta$ & $0.4630 \pm \Delta$\\
%  & VSAE (ours)  &$\bf{0.382\pm\Delta}$ & $\bf{0.3734\pm\Delta}$\\
 \bottomrule
\end{tabular}}
\end{table} 

%\vspace{-5mm}
Table~\ref{table:uci_mcar} shows that VSAE outperforms other methods on all datasets under MCAR missing mechanism in the missing ratio of 0.5. 
%VSAE can achieve lower imputation error with lower variance.
%The first five rows are trained in partially-observed setting, while the last two trained with fully-observed data. We observe that models trained with partially-observed data can outperform those models trained with fully-observed data on some datasets. We argue this is due to two potential reasons: (1) the mask provides a natural way of dropout on the data space, thereby, helping the model to generalize; (2) if the data is noisy or has outliers (which is common in low-dimensional data), learning from partially-observed data can improve performance by ignoring these data. However, although our model does not product state-of-the-art results in fully-observed data imputation settings, these models potentially can serve as upper bound if the data is clean. 
Fig.~\ref{fig:uci_ratios} illustrates VSAE generally achieves lower error along with relatively lower variance in all missing ratios. 
When the missing ratio increases (i.e.\ more data attributes become unobserved), VSAE consistantly maintains stable performance on most of the datasets. Conversely, we can observe a performance drop along with higher variance in the case of baselines. 
As the missing ratio increases, the attributive proposal network of VSAE maintains the same input, while the encoders of other methods have to learn to focus on the valuable information.
We believe the selection mechanism of proposal distribution in VSAE mitigates this negative effect.  %Unimodal encoders only take single observed modality as input and ignore the noisy unobserved information. 

 \begin{table*}[t]
%\vspace{-1.5em}
%\footnotesize
\centering
\caption{\textbf{Data Imputation on multi-modal datasets.} Missing ratio is $0.5$. We evaluate each dataset w.r.t.\ each attribute---label attribute is evaluated by \textbf{PFC}, image attributes of MNIST and FashionMNIST are evaluated by \textbf{MSE} averaged over pixels, other attributes are evaluated by \textbf{MSE}. Lower is better for all. We show mean and standard deviation over 3 independent trials. $\Delta < 0.001$. }
%\label{tab:uci_results}
\scalebox{0.8}{\begin{tabular}{lcccccccccc}
\toprule
& \multicolumn{2}{c}{\textbf{FashionMNIST + label (PFC)}} && \multicolumn{2}{c}{\textbf{MNIST + MNIST}}&& \multicolumn{3}{c}{\textbf{CMU-MOSI}}  \\
\cmidrule{1-3}\cmidrule{5-6}\cmidrule{8-10}
 {Attribute type}&Image & Label && Digit--1 & Digit--2 && Text & Audio & Image \\
\midrule
AE 
& $0.1105\pm0.001$ & $0.366\pm0.01$& 
& $0.1077\pm\Delta$ & $0.1070\pm\Delta$ &
&$0.035\pm0.003$ & $0.224\pm0.025$ & $0.019\pm0.003$\\
VAE
& $0.0885\pm\Delta$&$0.411\pm0.01$ &
& $0.0734\pm\Delta$ & $0.0682\pm\Delta$ & 
&$0.034\pm\Delta$ & $0.202\pm0.003$ &  $\bf{0.017\pm\Delta}$\\
CVAE w/ mask 
& $0.0887\pm\Delta$ & $0.412\pm0.01$& 
& $0.0733\pm\Delta$ & $0.0679\pm\Delta$ &  
&$0.043\pm\Delta$ & $0.257\pm0.002$ &  $0.020\pm\Delta$ \\
MVAE 
& $0.1402\pm0.026$ & $0.374\pm0.07$& 
& $0.0760\pm\Delta$ & $0.0802\pm\Delta$ &  
& $0.44\pm\Delta$ & $0.213\pm0.001$ & $0.025 \pm \Delta$ \\
HI-VAE 
& $0.1575\pm0.006$ & $0.405\pm0.01$& 
& $0.0772\pm\Delta$ & $0.0725\pm\Delta$ &  
& $0.047\pm\Delta$ & $0.211\pm0.005$ &  $ 0.0267\pm\Delta$ \\ 
VSAE (ours)
& $\bf{0.0874\pm\Delta}$ & $\bf{0.356\pm0.01}$& 
& $\bf{0.0712\pm\Delta}$ & $\bf{0.0663\pm\Delta}$ &  
&  \color{black}$\bf{0.033\pm\Delta}$ &  \color{black}$\bf{0.200\pm\Delta}$ &  \color{black}$\bf{0.017\pm\Delta}$ \\ 

\bottomrule
\end{tabular}}
%\vspace{2em}
\label{mnist_mnist_svhn_0.5}
\end{table*}

\begin{figure*}[t]
\begin{floatrow}
\ffigbox[\FBwidth]{%
 \includegraphics[width=0.985\linewidth]{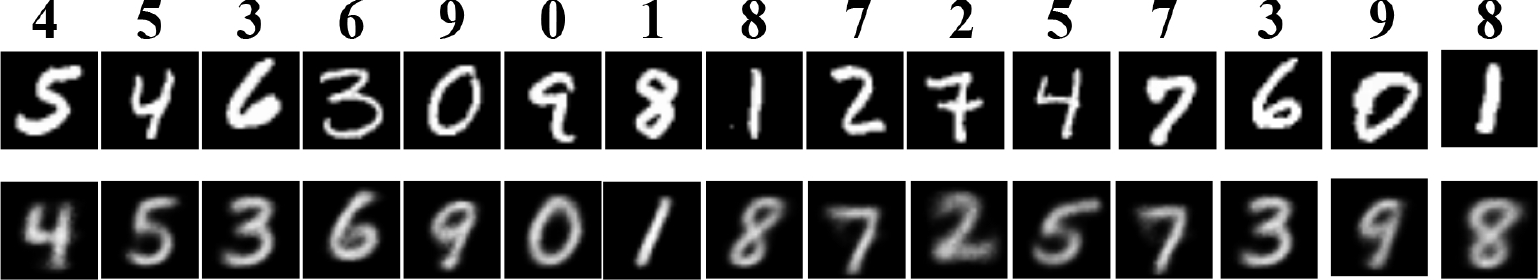}
}{%
  \caption{\textbf{Data Imputation on MNIST+MNIST.} Top is the labels of unobserved digit via pre-defined rules; middle is the observed attribute; bottom shows the imputation of unobserved attribute from VSAE.}
  \label{fig:mm_imp}
 }
\ffigbox[\FBwidth]{%
  \includegraphics[width=0.94\linewidth]{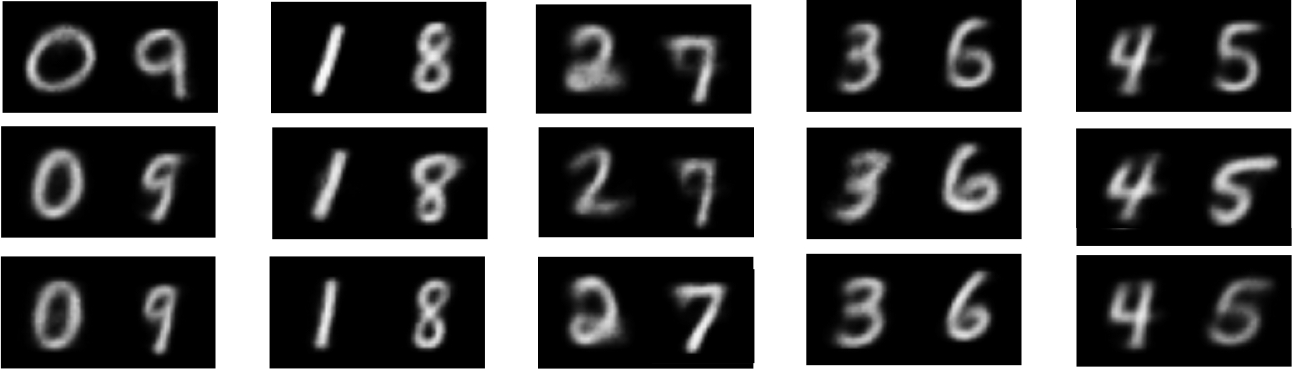}
}{%
  \caption{\textbf{Data Generation on MNIST+MNIST.} Images are generated without conditional information. As shown, the correspondence between attributes are preserved in stochastic data generation. }
  \label{fig:mm_gen}
}
\end{floatrow}
\end{figure*}

In Fig.~\ref{fig:uci_ratios}, a seemingly strange behaviour is observed --- some models  (e.g. HI-VAE for \textit{Phishing}) perform better with more missing data. We hypothesize that: 
{(i)} The missingness provides a natural way of dropout on the raw data space.
% For some thereby under certain missing ratios possibly helping some model to generalize;
Under certain missing ratios, it possibly reduces overfitting. Hence, it can help the model to generalize;
{(ii)} To synthesize the missing mechanism, we sample and fix a mask on each dataset for all models, but the model itself may also be sensitive to the mask randomness.
% More missing data potentially increases the chance to ignore some corrupted data or outliers.
Under a certain random sampling of the mask, some corrupted data or outliers, where one model is sensitive, might be observed and degrade the performance of this model.
For example, on \textit{Phishing} dataset, HI-VAE performs slightly better at 0.5 than 0.3). The error bar is wider at 0.3, but relatively narrow at 0.5 and 0.7. 
Some outliers observed at 0.3 may contribute to this variance.
% The variance at 0.3 missing ratio may be caused by observed outliers.
It performs worse again at 0.7, as the positive effect by masking out the outliers is not dominant when the missing ratio is too high.

%%%%%%%%%%% Non-MCAR

\textbf{\textit{Non-MCAR masking.}} 
%VSAE jointly models data and mask distribution without assumption on mask distribution. 
MIWAE~\citep{mattei2019miwae} assumes data are MAR and approximates the conditional expectation of the unobserved attributes using importance sampling. 
%~\cite{mattei2019miwae} conducted experiments with synthesized mask in a MAR manner. 
We follow it to mimic non-MCAR missing mechanisms on UCI numerical datasets --- {MAR}: 25\% attributes are default observed, then sample the remaining mask from $\operatorname{sigmoid}(\frac{1}{M}\sum_{k=1}^K \x_k)$,
where $K$ is the number of observed attributes; {NMAR}: sample the mask for $\x_i$ from $\operatorname{sigmoid}(\x_i)$. We use the public code of MIWAE and the same imputation estimates (single imputation of 1000 importance samplings) for fair comparison.
Table~\ref{table:uci_nonmcar} indicates VSAE can outperform the state-of-the-art non-MCAR model MIWAE in the non-MCAR setting, particularly for the NMAR, as VSAE models the joint distribution of attributes and mask without introducing the presumably false independence among them.

% \textit{Missing At Random (MAR):} The mask solely depends on the observed attributes. We choose 25\% attributes as default observed data ($m_i=1$), then sample the mask of remaining attributes from probability 
% $\operatorname{sigmoid}(\frac{1}{M}\sum_{k=1}^K \x_k)
% $,
% where $M$ and  $K$ is the number of attributes and default observed
% attributes.

% \textit{Not Missing At Random (NMAR):} The mask distribution depends on observed and unobserved attributes. We sample the mask from the probability $
%     \pi(m_i)=\operatorname{sigmoid}(\x_i)$, where
% $m_i$ is $i$-th element of mask $\mathbf{m}$, $\x_i$ is the $i$-th attribute. 

% \begin{table}[t]
% \caption{\textbf{Non-MCAR Data Imputation.} Missing mechanism is defined as above. We show mean and standard deviation of NRMSE over 3 independent runs, lower is better. $\Delta<0.01$. }\label{table:uci_nonmcar}
%  \scalebox{0.87}{
% \begin{tabular}{c c  c c} 
%  \toprule
% &\textbf{Method}  &  MAR & NMAR \\ %[0.5ex] 
% \midrule
%   \multirow{2}{*}{\textbf{Yeast}} 
%   & MIWAE & $0.493\pm0.025$ & $0.513 \pm 0.036$\\
%   & VSAE (ours)  & $\bf{0.472\pm0.016}$ & $\bf{0.456\pm\Delta}$\\
%  %NMAR 
%  \midrule
%  \multirow{2}{*}{\textbf{Whitewine}} 
%  & MIWAE & $0.493\pm\Delta$ & $0.463 \pm \Delta$\\
%  & VSAE (ours)  &$\bf{0.382\pm\Delta}$ & $\bf{0.373\pm\Delta}$\\
% %   \multirow{2}{*}{\textbf{Whitewine}} 
% %  & MIWAE & $0.493\pm\Delta$ & $0.4630 \pm \Delta$\\
% %  & VSAE (ours)  &$\bf{0.382\pm\Delta}$ & $\bf{0.3734\pm\Delta}$\\
%  \bottomrule
% \end{tabular}}
% %\end{center}
% \end{table}

\paragraph{Mask Generation.}
% \begin{wraptable}{r}{.4635\columnwidth}
% \caption{\textbf{Mask Generation.} The average proportion of 0 is calculated on 100 sampled mask variables, averaged over all experiments.} \label{table:mask_gen}
% \scalebox{0.8}{
% \begin{tabular}{c c} 
%  \toprule
%  \textbf{Miss. ratio} & \textbf{Avg. prop. of 0}\\
%  \midrule
%  $0.3$  & $0.312\pm0.016$ \\
%   $ 0.5$ & $0.496\pm0.009$ \\ 
%   $ 0.7$ & $0.692\pm0.005$ \\
%  \bottomrule
% \end{tabular}}
% \end{wraptable} 
VSAE enables us to generate data and mask from the learned generative model $p_{\boldsymbol{\theta},\boldsymbol{\epsilon}}(\x,\mathbf{m},\mathbf{z})$. 
We show mask generation results on UCI and data generation on multi-modal datasets (Sec.~\ref{sec:gen_multimodal}), since the sole data generation is not qualitatively or quantitatively measurable in the case of UCI datasets. 
%The mask conditional log-likelihood term in Eq.~\eqref{eq:elbo_final} allows the latent representation embedded with mask information and thus can reconstruct (or generate if sampling from the prior) the mask variable. 
%In the \textbf{MCAR} setting, the mask follows a Bernoulli distribution governed by the predefined missing ratio. 
After training, we can sample from the prior to generate the mask on the corresponding dataset.

We evaluate mask generation by calculating the average proportion of missing attributes ($m_i=0$) on 100 sampled masks. In the \textbf{MCAR} setting, averaged over datasets, we get 0.312$\pm$0.016, 0.496$\pm$0.009, 0.692$\pm$0.005 for the ground truth missing ratios of 0.3, 0.5, 0.7. In the \textbf{MAR} setting where the mask depends on the dataset, we get 0.631$\pm$0.026 and 0.665$\pm$0.019 for the ground truth missing ratios of 0.626 (Whitewine) and 0.676 (Yeast). Similarly, in the \textbf{NMAR} setting we get 0.513$\pm$0.021 and 0.592$\pm$0.013 for the ground truth missing ratios of 0.502 (Whitewine) and 0.583 (Yeast). It indicates VSAE is capable of accurately learning the mask distribution.
% 0.631$\pm$0.026 /  for the ground truth missing ratio of 0.6755(MAR) / 0.5830(NMAR) on Yeast.  
% / 0.513$\pm$0.021
% (MAR) / 0.502(NMAR) 
Interestingly, we find improvement in the performance by conditioning the reconstructed mask variable on the data decoders. We speculate that this may be because the learned distributional mask variable can inform the data decoder of the missingness distributed in the data space, which in turn allows the potential missing mechanism to guide the data generative process.

\begin{table*}[t]
\centering
\caption{\textbf{Ablation study on modules of VSAE.} We consider experiments on \textit{Yeast} and \textit{Whitewine}. The masks of MCAR, MAR and NMAR are synthesized by the rules mentioned in Sec.~\ref{sec::tabular_data}.  We evaluate each dataset by \textbf{NRMSE}. Lower is better. We show mean and standard deviation over 3 independent trials. $\Delta<0.0005$.}
\label{table:ablation}
\scalebox{0.83}{\begin{tabular}{lcccccccc}
\toprule
%\cmidrule{2-3}\cmidrule{5-6}
 & \multicolumn{2}{c}{\textbf{MCAR}} & &\multicolumn{2}{c}{\textbf{MAR}} &&  \multicolumn{2}{c}{\textbf{NMAR}} \\
 \cmidrule{2-3}\cmidrule{5-6}\cmidrule{8-9}
 & \textbf{Yeast}&\textbf{Whitewine} & &\textbf{Yeast}&\textbf{Whitewine} &&  \textbf{Yeast}&\textbf{Whitewine} \\
\midrule

 {Missing ratio}
& 0.5 & 0.5 & 
& 0.676 & 0.626 &
& 0.583 & 0.502 \\

%\cmidrule{1-3}\cmidrule{5-6}\cmidrule{8-9}
 %\textbf{Attribute type} &categorical&categorical  &&numerical&numerical &&categorical&numerical \\
\midrule

 \textcolor{black}{VSAE (collective only)}
& ${0.426\pm0.001}$ & ${0.3730\pm\Delta}$ & 
& ${0.480\pm\Delta}$ & ${0.3837\pm\Delta}$&
& $\bf{0.427\pm0.005}$ & ${0.3737\pm\Delta}$ \\

\textcolor{black}{VSAE (attributive only)}
&  ${0.457\pm0.017}$ & ${0.3728\pm\Delta}$ & 
&  ${0.493\pm0.005}$ & ${0.3827\pm\Delta}$ &
&  ${0.434\pm0.006}$ & ${0.3729\pm\Delta}$ \\
  \midrule

\textcolor{black}{VSAE (w/o mask modeling)}
& $\bf{0.418\pm0.004}$ & ${0.3720\pm\Delta}$ & 
& ${0.485\pm0.006}$    & ${0.3834\pm\Delta}$ &
& ${0.437\pm\Delta}$   & ${0.3724\pm\Delta}$ \\
\midrule
No EM 
&  $\bf{0.426\pm0.012}$ & ${0.3723\pm\Delta}$ & 
&  ${0.476\pm0.001}$ & $\bf{0.3821\pm\Delta}$ &
&  ${0.432\pm0.003}$    & ${0.3727\pm\Delta}$ \\
EM with 1 sampling
& ${0.433\pm0.002}$     & $\bf{0.3719\pm\Delta}$ & 
& $\bf{0.473\pm\Delta}$ & ${0.3825\pm\Delta}$ & 
& $\bf{0.426\pm0.003}$     & ${0.3721\pm\Delta}$  \\

EM with 1000 samplings
& $\bf{0.418\pm0.008}$ & $\bf{0.3718\pm\Delta}$  & 
& $\bf{0.471\pm0.007}$ & ${0.3826\pm\Delta}$   & 
& $\bf{0.424\pm0.005}$ & $\bf{0.3716\pm\Delta}$  \\
 \midrule

VSAE (ours)
& $\bf{0.419\pm0.008}$ & $\bf{0.3719\pm\Delta}$  & 
& $\bf{0.472\pm0.006}$ & ${0.3825\pm\Delta}$ &
& $\bf{0.425\pm0.007}$ & $\bf{0.3717\pm\Delta}$\\
\bottomrule
\end{tabular}}
\end{table*}
% ************************ Multiu-Modal
\subsection{Multi-modal Data}

\cite{BaltrusaitisAM17} defined \textit{multi-modal data} as data of multiple \textit{modalities}, where each modality is a way to sense the world---seeing, hearing, feeling, etc. 
%In order to understand the world, it needs to interpret such multi-modal signals, by processing and relating information from multiple modalities.
However, our definition of \textit{multi-modal} covers a wider spectrum where the data could be of the same type (e.g.\ image) but from different distributions (e.g. different shapes). 
In the manner multi-modal data are collected or represented, we can safely treat multi-modal data (typically high-dimensional) as a type of heterogeneous data. In the following, we use \textit{attribute} and \textit{modality} interchangeably as a notion of heterogeneity.
We design experiments on three types of multi-modal data: 
(i) image/label pair--- \textbf{{FashionMNIST}} images and labels; 
(ii) image/image pair---synthesized bimodal MNIST+MNIST datasets by pairing two different digits from \textbf{{MNIST}} as \{(0,9),(1,8),(2,7),(3,6),(4,5)\}; 
(iii) standard multi-modal dataset \textbf{{CMU-MOSI}}~\citep{zadeh2018multi} including visual, textual and acoustic signals.  See Appendix E for more results.
For all we use the standard training/validation/test split and all masking follows MCAR. We evaluate the performance on labels with \textbf{{PFC}} (proportion of falsely classified attributes), images with \textbf{{MSE}} (mean-squared error) averaged over pixels and other attributes with \textbf{MSE}. %~\ref{sec:additional_results}. 
 %We also report experimental results on multimodal FashionMNIST, MNIST and CMU-MOSI. 

\paragraph{Data Imputation.}

% \begin{figure}{
% \includegraphics[width=1\linewidth]{figures/data_impt.pdf}
% }{%
%   \caption{\textbf{Data Imputation on MNIST+MNIST.} Top is the labels of unobserved digit via pre-defined rules; middle is the observed attribute; bottom shows the imputation of unobserved attribute from VSAE.}
%   \label{fig:mm_imp}}
% \end{figure}

Table~\ref{mnist_mnist_svhn_0.5} demonstrates VSAE can achieve superior performance for multi-modal data imputation on all modalities with lower variance. Fig.~\ref{fig:mm_imp} presents the qualitative results of imputations on MNIST+MNIST image pairs. 
To demonstrate robustness to missing ratio, we conducted experiments with missing ratio of 0.3, 0.5, 0.7 on the MNIST+MNIST dataset and the sum errors by VSAE are 0.1371$\pm$0.0001, 0.1376$\pm$0.0002 and 0.1379$\pm$0.0001 respectively. This indicates that VSAE also stays robust under different missing ratios for multi-modal datasets. 
%(see Appendix~\ref{subsec:mm_bimodal_mr} for more results). This indicates that VSAE also stays robust under different missing ratios for multi-modal datasets. 

%The multimodal encoders try to find the information needed to generate the unobserved modalities from the available observed information. 

%Multimodal encoders also include the mask vector as input. This allows the multimodal encoders to be aware of the shape of the missingness and forces it to focus on the useful information in the observed modalities.

\paragraph{Data Generation.}\label{sec:gen_multimodal}
%\subsubsection{Generation on synthetic dataset}\label{sec:gen_multimodal}

% \begin{figure*}[t]
% \begin{floatrow}
% \ffigbox[\FBwidth]{%
%  \includegraphics[width=1\linewidth]{figures/data_impt.pdf}
% }{%
%   \caption{\textbf{Data Imputation on MNIST+MNIST.} Middle is the observed attribute with labels of unobserved digit given in the top  via pre-defined rules. Bottom shows the imputation of unobserved attribute from VSAE.}
%   \label{fig:mm_imp}
%  }
% \ffigbox[\FBwidth]{%
%   \includegraphics[width=0.96\linewidth]{figures/data_gen.pdf}
% }{%
%   \caption{\textbf{Data Generation on MNIST+MNIST.} Images are generated w/o conditional information. As shown, the correspondence between attributes are preserved. }
%   \label{fig:mm_gen}
% }
% \end{floatrow}
% \end{figure*}

Fig.~\ref{fig:mm_gen} shows that VSAE is capable of generating image-based attributes following the underlying correlation.
The learning process does not require any supervision and can be effectively carried out with only access to partially-observed data.
% Interestingly, we find improvement on the performance by conditioning the reconstructed mask variable on the data decoders. We speculate that this may be because the mask variable can inform the data decoder of the missingness distributed in the data space, which in turn allows the potential missing mechanism to guide the data generative process.
% \begin{figure}
% {%
%  \includegraphics[width=0.97\linewidth]{figures/data_gen.pdf}
%   \caption{\textbf{Data Generation on MNIST+MNIST.} Images are generated w/o conditional information. The correspondence between attributes are preserved. }
%   \label{fig:mm_gen}
% }\end{figure}
We believe the underlying mechanism of selective proposal distribution benefits the performance. Though unobserved attributes in one data sample are not available even during training, they could be the observed attributes in others. 
Thus, the collective proposal networks are able to construct a stochastic mapping from observable to unobservable information among the whole training set. 
The separate structure of {attributive} and {collective proposal network} enforces VSAE to attend to the observed attributes, by ignoring unobserved attributes and performing efficient heterogeneity aggregation in the latent space. Thus it shows consistent robustness to various missing ratios. 
In contrast, baselines primarily approximate the posterior by a single proposal distribution inferred straight from the whole input. 
%VSAE readily ignores noisy unobserved attributes and attends on useful observed attributes, while 
As a result baselines rely heavily on neural networks to extract expressive information from the data, which can be dominated by missing information.

\subsection{Ablation Study}
In Table~\ref{table:ablation}, we perform extensive ablation studies to demonstrate the effectiveness of three critical modules: selective proposal network, mask distribution modeling, and expectation over generated $\x_\uu$ during training.

\textbf{Selective proposal distribution.} We compare VSAE to its variants with only collective or only attributive proposal network, which shows both are essential to the improvement. 
Empirically we observe the collective structure tends to outperform its attributive counterpart on \textit{Yeast} but reversely worse on \textit{Whitewine}. In fact, if we analyze the datasets, \textit{Yeast} has highly correlated attributes (e.g. scores from different methods for the same task), whereas \textit{Whitewine} has more independent attributes (e.g. density, acidity or sugar). 

\textbf{Modeling the mask.} We exclude mask conditional log-likelihood and make data decoder conditioned on the input mask. We observe the performance drops for non-MCAR, and do not observe significant improvement for MCAR. It indicates the mask reconstruction does not sacrifice model capacity in the case of MCAR, where the mask distribution can be easily learned.

\textbf{Expectation over generated $\x_\uu$.} This learning procedure mitigates the intractability of $\x_\uu$, therefore allowing to model the dependence between $\x_\uu$ and $\mm$. We first consider removing this expectation procedure and ignoring $\x_\uu$ during training. The result (No EM) indicates this procedure plays an important role in NMAR ($\x_\uu$ depends on $\mm$) and does not significantly sacrifice the performance of MCAR and MAR. Compared to larger sampling size (of 1000), the result shows our proposed method (which generates 100 samples of $\x_\uu$ to take expectation during training) is sufficient to obtain a good estimation of the expectation.
\section{Conclusion}
\label{sec:conclusions}

In this work, we propose VSAE, a novel latent variable model to learn from partially-observed heterogeneous data. The proposed VSAE handles  missingness effectively by introducing a selective proposal distribution which is factorized w.r.t the data attributes. Further, VSAE is a general framework and capable of performing multiple tasks including data imputation, data generation and mask generation. 
Extensive experiments with comparison to state-of-the-art deep latent variable models demonstrated the effectiveness of VSAE on a variety of tasks. 
We summarize our contributions in the partially-observed heterogeneous setting as follows: 

\textbf{{{Heterogeneity.}}} The {factorization w.r.t. attributes in the latent space} reduces the negative impact from the heterogeneity in the raw data space.

\textbf{{{Partial observations.}}} VSAE approximates the true posterior given partially-observed data with a novel {selective proposal distribution}. The automatic encoder selection between observed and unobserved attributes enables VSAE to ignore noisy information
%from unobserved attributes 
and learn from partial observations. 

\textbf{{{No MCAR assumption.}}} The independence assumption between data and mask can be restrictive. VSAE relaxes this assumption and models the joint distribution of data and mask together. To handle the intractablility of $\x_\uu$, an expected lower bound is maximized during training.

\bibliographystyle{plainnat} 
\bibliography{vsae}

\end{document}